\documentclass{article}

\usepackage[final]{neurips_2019}

\usepackage[utf8]{inputenc} 
\usepackage[T1]{fontenc}    
\usepackage{hyperref}       
\usepackage{url}            
\usepackage{booktabs}       
\usepackage{amsfonts}       
\usepackage{nicefrac}       
\usepackage{microtype}      
\usepackage{graphicx}

\usepackage{comment}
\usepackage{authblk}

\author[1]{ \textbf{Caleb Hoyne}}
\author[1,2,*]{ \textbf{S. Karthik Mukkavilli}}
\author[1]{  \textbf{David Meger}}

\affil[1]{McGill University}
\affil[2]{Mila - Quebec AI Institute}
\affil[*]{Corresponding author:\{karthik\}@cim.mcgill.ca}
\begin{document}
\title{Deep learning for Aerosol Forecasting}
\maketitle
\begin{abstract}

Reanalysis datasets combining numerical physics models and limited observations to generate a synthesised estimate of variables in an Earth system, are prone to biases against ground truth. Biases identified with the NASA Modern-Era Retrospective Analysis for Research and Applications, Version 2 (MERRA-2) aerosol optical depth (AOD) dataset, against the Aerosol Robotic Network (AERONET) ground measurements in previous studies, motivated the development of a deep learning based AOD prediction model globally. This study combines a convolutional neural network (CNN) with MERRA-2, tested against all AERONET sites. The new hybrid CNN-based model provides better estimates validated versus AERONET ground truth, than only using MERRA-2 reanalysis.

\end{abstract}
    
\section{Introduction and Related Work}

Machine learning can help tackle challenges in climate models \citep{Rolnick}, where atmospheric aerosols are a large source of uncertainty. Ground truth aerosol (atmospheric particulates) data is not readily available in all locations. Reanalysis datasets such as NASA Modern-Era Retrospective Analysis for Research and Applications, Version 2 (MERRA-2) provide an estimation of AOD using satellite data and physics models, amongst other sources. Because ground truth AOD values are not available at all sites, MERRA-2 can inform AOD predictions where there are no aerosol measurements. However, potential error in characterizing AOD at locations due to the relatively low spatial resolution of satellite data compared to ground truth and issues with retrieval algorithms from satellite imagery can introduce biases in reanalysis as in \cite{mukkavilli}.

Prior work focused on the usage of neural networks in this domain that have primarily relied on simple single or multilayer perceptron (MLP) architectures. The work of \cite{Malakar2012} examines the integration of MODIS AOD as input to a artificial neural network (ANN) to more accurately forecast AOD. This study used a MLP whose input was the vector of variables (e.g. solar zenith angle, AOD at different wavelengths etc) hypothesized to be relevant in the bias of MODIS in addition to AOD. However, in this study, the network is a deep convolutional neural network (CNN), whose input is spatial imagery from reanalysis data without additional variables as input. studied the effect of an ensemble of ANNs in a stacked generalization approach with adaptive cost functions on the performance of predictions. In the ensemble, some ANNs focused on large or extreme AOD values while others focused on more normal or background AOD observations. In addition, instead of using a mean squared error (MSE) loss function, a relative error function was used which weighted error relative to the magnitude of the observation to ensure it was equally important to predict low and high AOD values accurately. This study found that this ensemble of ANNs trained on a relative error loss function improved performance compared to that of a single ANN. Like the work of \cite{Radosavljevic2007}, two neural networks are trained in this study for normal and large AOD observations. However, they are not used in conjunction to maximize the accuracy of predictions. Another difference is that the input and output of the stacked generalization approach was a vector of AOD values that had a corresponding vector of target values as opposed to using a CNN with gridded spatial reanalysis data as input. \cite{Kleynhans2017} used a CNN to predict top-of-atmosphere thermal infrared irradiance with MERRA-2. It is worth considering CNNs for the use case of AOD as the properties of aerosols compared to other atmospheric variables are different. This study to the author's knowledge, is the first use of deep neural networks to forecast aerosols with reanalysis.

\section{Data}

The datasets used in this analysis are MERRA-2 and AERONET. MERRA-2 provides estimated AOD based on global reanalysis data as gridded spatial imagery (see example day in Fig. 1), whereas AERONET provides ground truth observations of AOD at various observation centers (Fig. 2). They will be analysed for a decade, spanning from 2008 to 2017.

\subsection{Reanalysis Physics Model and Dataset (MERRA-2)}

MERRA-2 AOD at 550nm is chosen because it is the peak wavelength of the visible spectrum. The spatial resolution of the dataset is 0.5 x 0.625. This low resolution creates challenges for accurately estimating AOD at AERONET sites that are furthest from the provided grid points. The temporal resolution is daily, where AOD is calculated as the daily mean of AOD recorded at 3-hour intervals. This data is available from 1980 to present day.

\subsection{Ground Truth Dataset (AERONET)} 

The AERONET project provides ground truth aerosol data from observation centers at several sites across globally. AOD at 500 nm is available as well as the Angstrom exponent for the range of 440nm-675nm. AOD at 550nm is estimated as in \cite{mukkavilli}. AERONET (version 3.0, level 2.0) AOD values are taken as the daily mean of the recorded data. This data is available 1993 to present day.

\begin{figure}[h]
 \centering
a\includegraphics[width=0.3\textwidth]{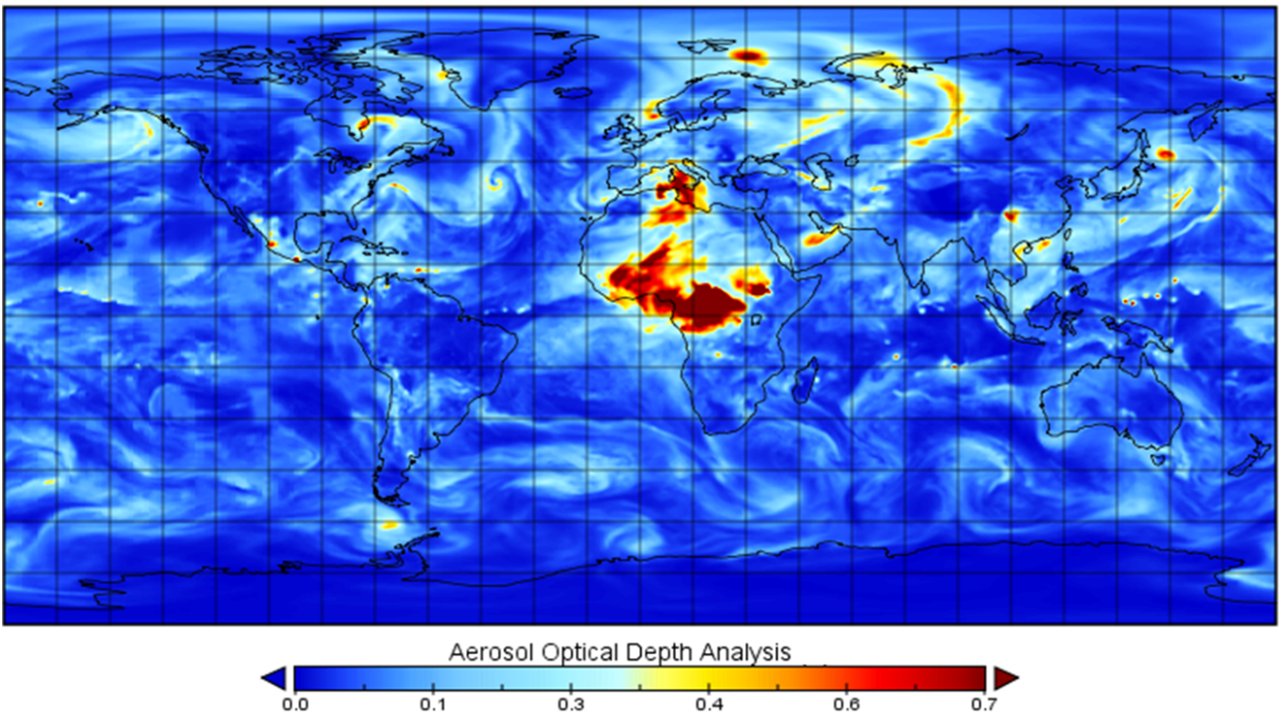}
b\includegraphics[width=0.3\textwidth]{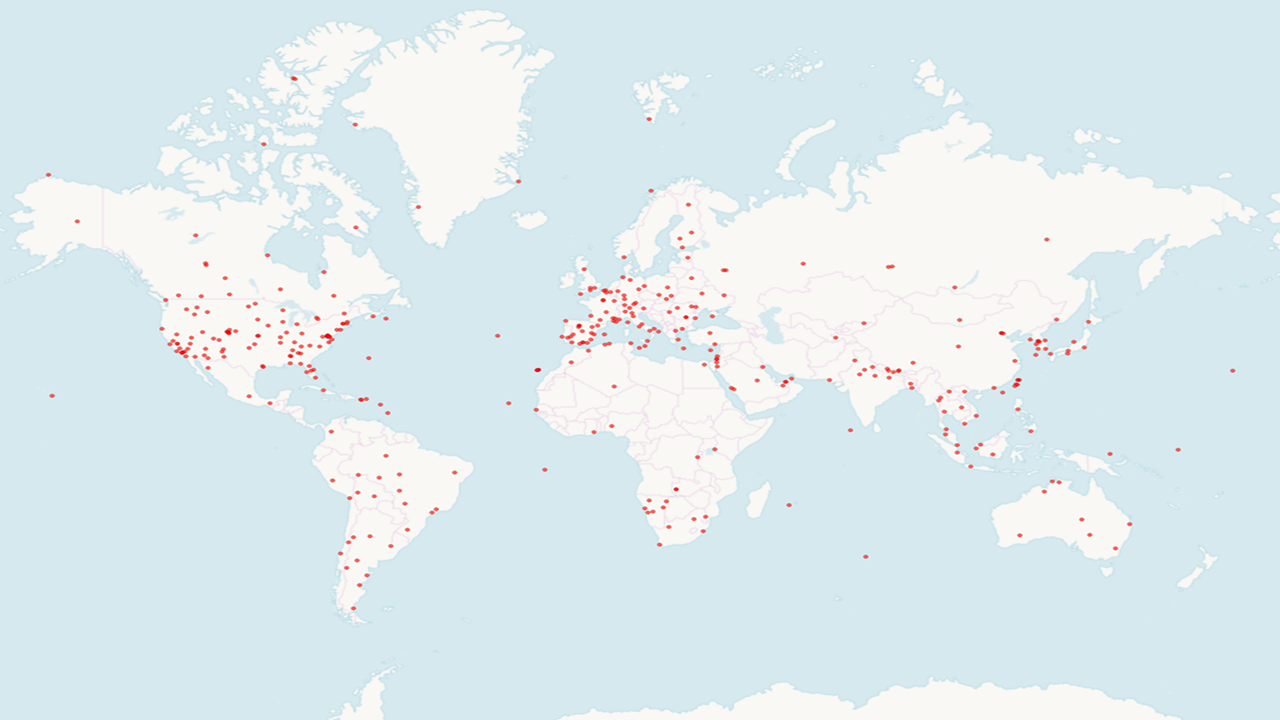}
      \caption{(a) Example daily mean MERRA-2 Total AOD global dataset (min = 0, mean = 0.1, max = 3.1) at 0.5\textsuperscript{o} x 0.625\textsuperscript{o} with bicubic interpolation (b) Location of AERONET sites globally}  
\end{figure}

\section{Results and Discussion}
\subsection{Global Comparison between MERRA-2 physics and AERONET truth}

Due to the difference in spatial resolution between the two datasets, AERONET observations are compared to the nearest grid point of the MERRA-2 dataset. This comparison yields a maximum distance between observations of approximately 40km. The comparisons are made at the same time step for each dataset to evaluate the error metrics of the MERRA-2 dataset against ground truth AOD. 

Between 2008-2017 a total of 464,661 observations included 17, 089 extreme observations (total AOD > 0.7). A general comparison of MERRA-2 and AERONET shows that MERRA-2 underpredicts AOD globally and particularly during extreme cases (Table 1). Extreme events incur a significantly greater error than normal days. The MBE for normal days is -0.041 and during extreme events it is -0.984. The RMSE of extreme events is nearly 7 times greater than that of normal days. Likewise, the MBE and MAE for extreme events is around 25 times greater than the respective statistics for normal days (Table 1).

\begin{table} [h]
  \caption{General comparison of MERRA-2 and AERONET dataset error metrics globally (2008-2017) for normal days, extreme events and all data}
  \label{table:lst}
  \centering
  \begin{tabular}{llll}
    \toprule
    Type of AOD Days &  RMSE & MAE & MBE \\
    \midrule
Normal & 0.166 & 0.038 & -0.041 \\
Extreme & 1.098 & 0.986 & -0.984 \\
All & 0.266 & 0.141 & -0.076 \\
    \bottomrule
  \end{tabular}
\end{table}

RMSE, MBE and MAE is greatest in the southeast Asia region, especially in India, China and Indonesia (Figure 2). The RMSE observed at these sites ranges from 0.707 to 0.989 which is substantially greater than the global RMSE of 0.2661. At these sites, the MBE is nearly always negative and similar in absolute value to its respective MAE. This suggests that for the sites which experience the most error, MERRA-2 consistently under-predicts AOD. Palangka Raya, Indonesia, the site with the highest variance, recorded the highest RMSE across extreme events, 2.404 and the site with the second highest variance, Jambi, Indonesia, recorded the second highest RMSE across extreme events, 1.57. This is substantially higher than the global average RMSE for extreme events of 1.097.  

\begin{figure}[h]
  \centering
\includegraphics[width=0.3\textwidth]{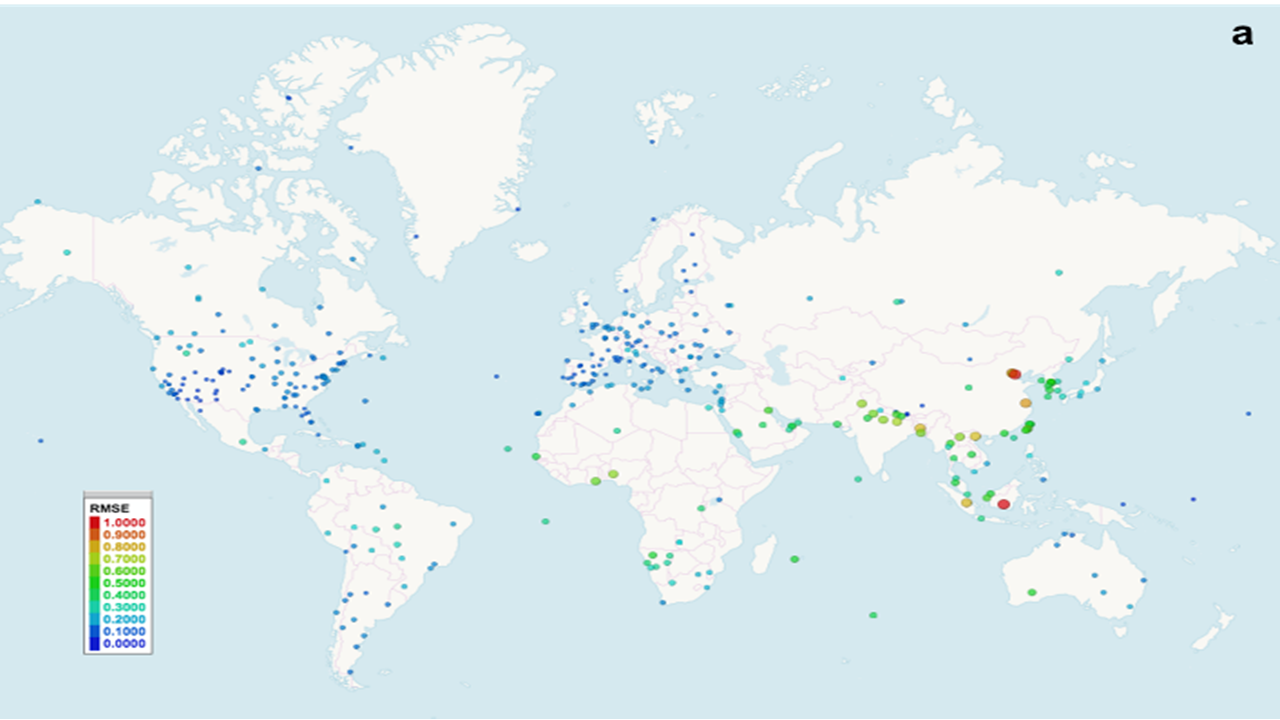}
\includegraphics[width=0.3\textwidth]{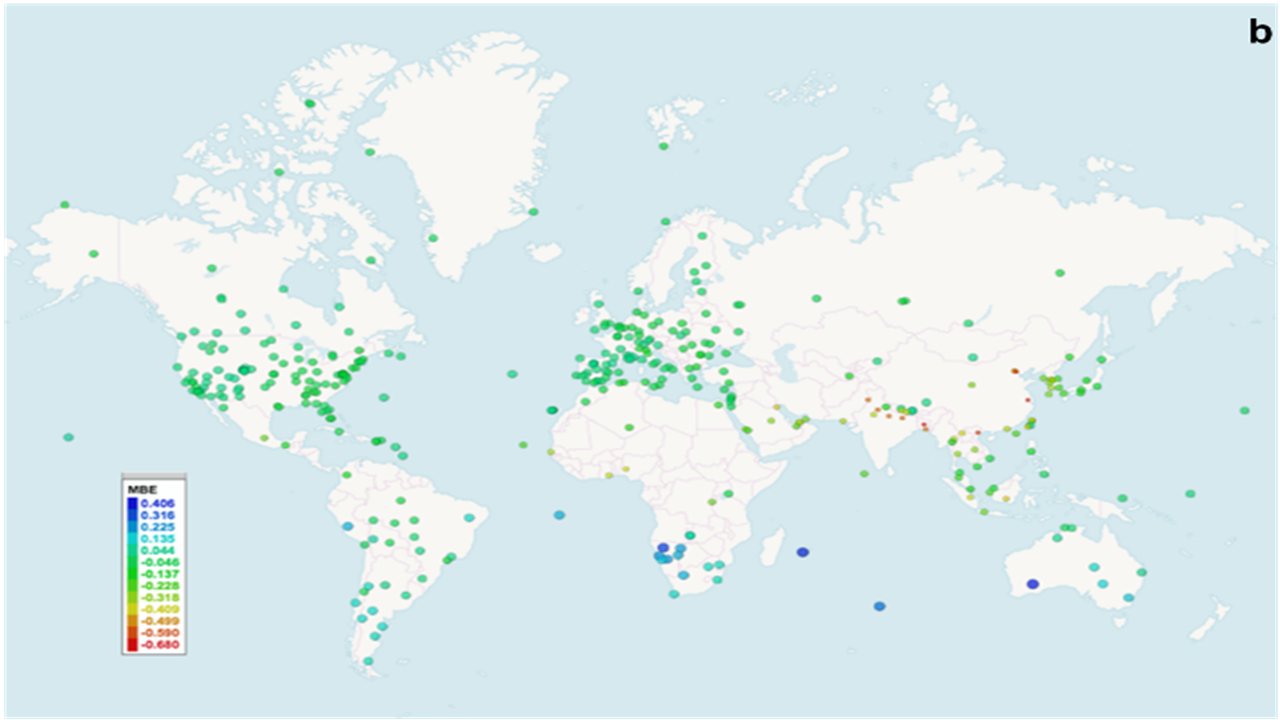}
\includegraphics[width=0.3\textwidth]{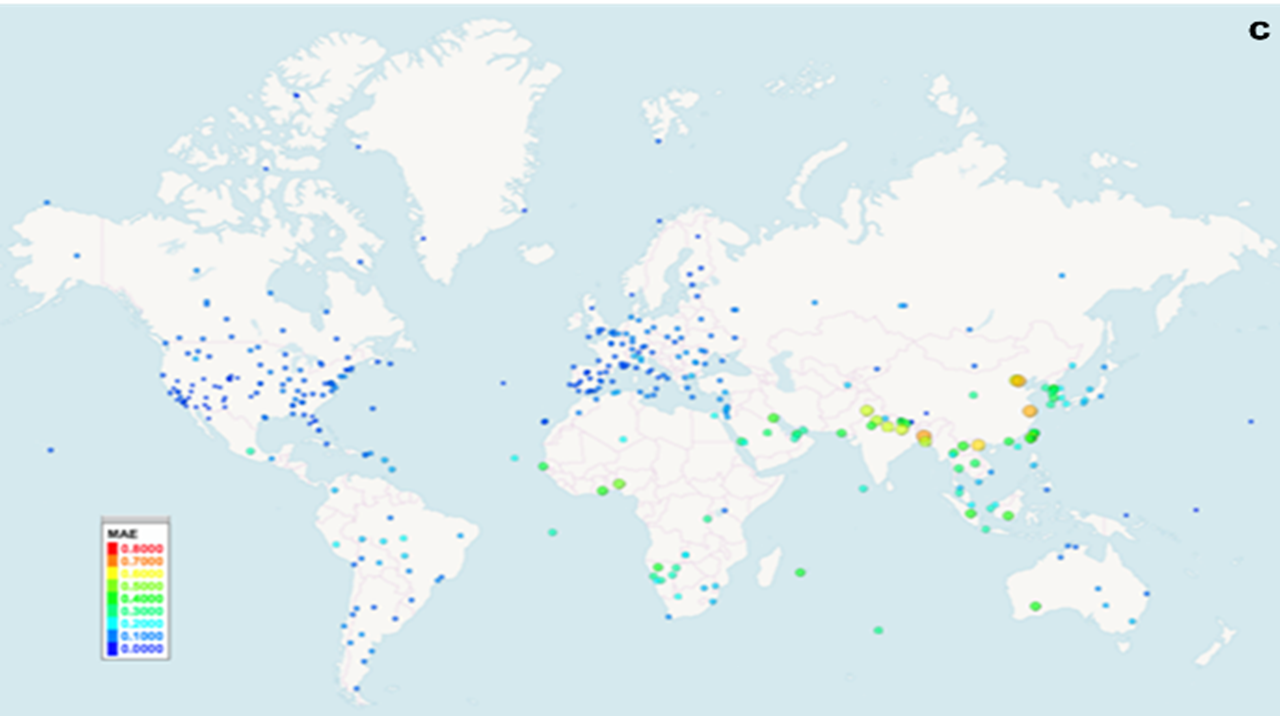}
      \caption{MERRA-2 error statistics against AERONET ground truth sites (with >365 observations) (a) RMSE (b) MBE (c) MAE}  
\end{figure}

The sites in which the most extreme observations occur include locations in India, China, Pakistan, and Nigeria. These locations all have high errors across all statistics. These results suggest that an improvement of AOD predictions is most useful for extreme events. The sites in which the most extreme observations occur are in Asia and Africa . The top five sites with extreme AOD include sites in the countries India, China, Pakistan, Nigeria, and Bangaladesh. The top five sites with high variance of AOD were all in Asia. 
  
\begin{figure}[h]
\centering
\includegraphics[width=0.3\textwidth]{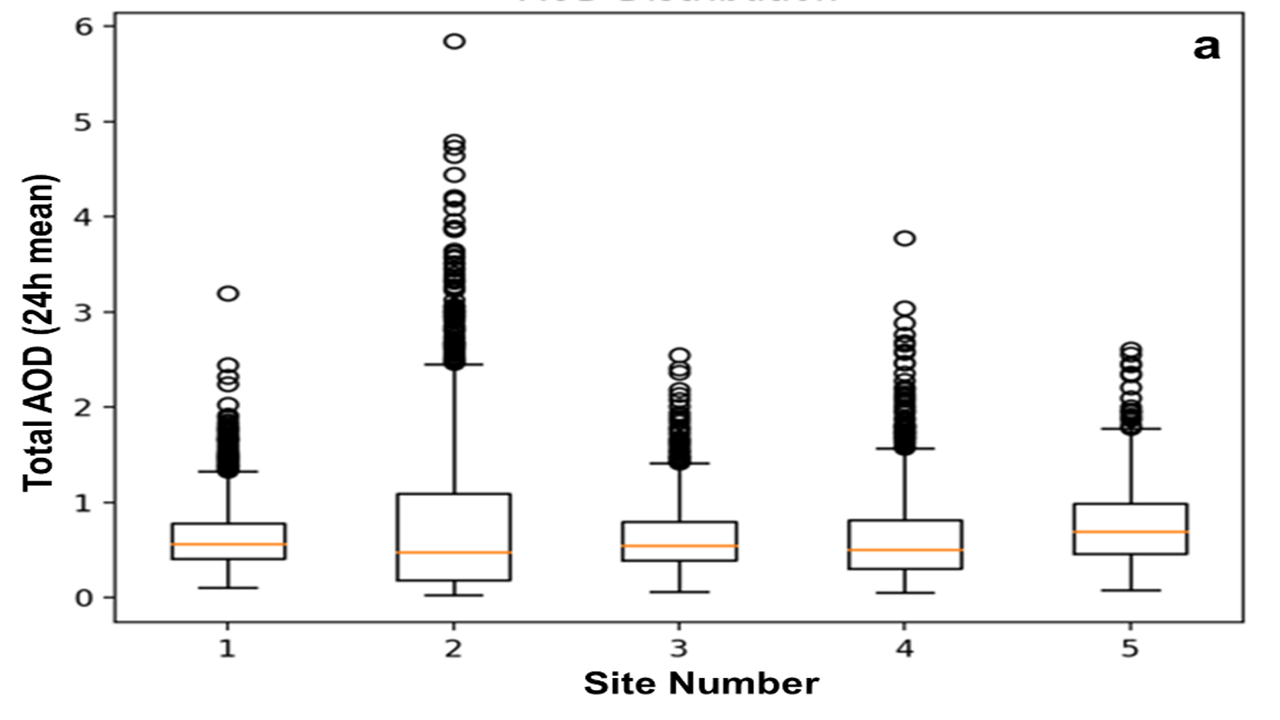}
\includegraphics[width=0.3\textwidth]{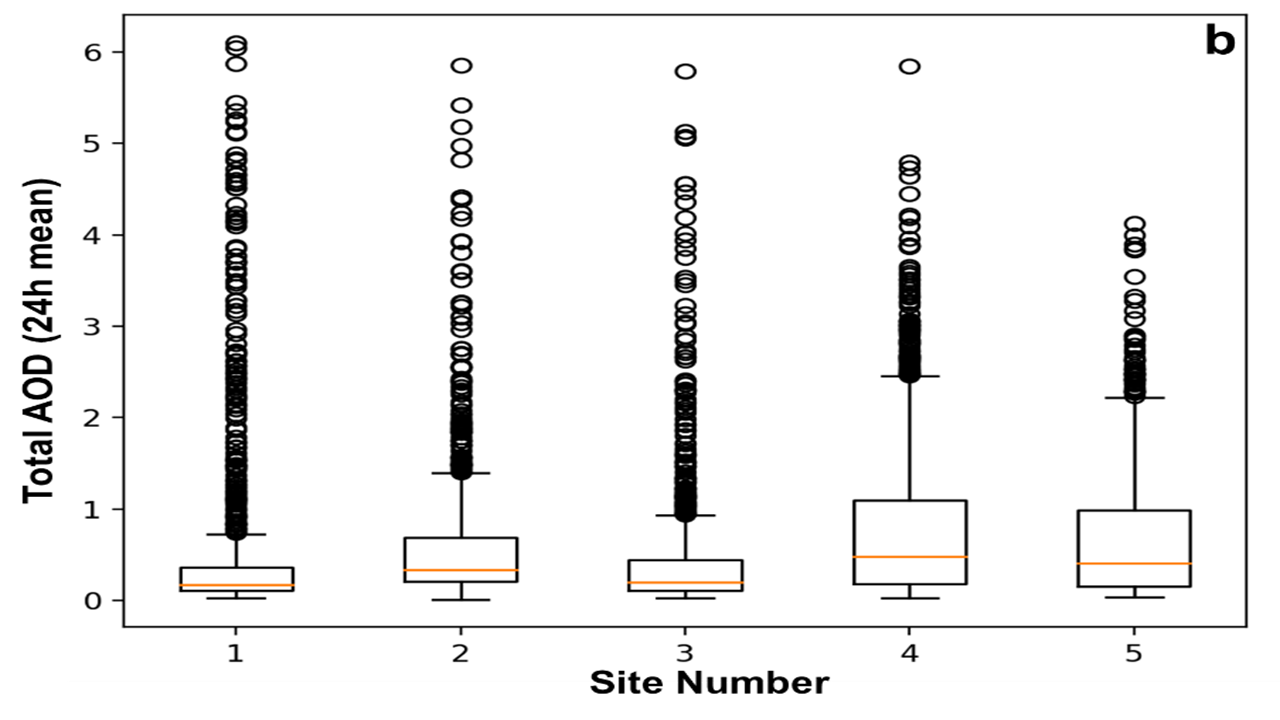} \caption{(a) Sites with the highest number of extreme AOD events: 1. Kanpur, India; 2. Xianghe, China; 3. Lahore, Pakistan; 4. Ilorin, Nigeria; 5. Dhaka Univ., Bangladesh (b) Sites with the highest AOD variance: 1. Palangka Raya, Indonesia; 2. Jambi, Indonesia; 3. Pontianak, Indonesia; 4. Xianghe, China; 5. Beijing, China}  
\end{figure}  

The sites where the most extreme observations occur tend to be in highly polluted areas. These areas also exhibit a high variance between extreme events to non-extreme events. Indonesia seems to have a particularly high variance which could be due to the distinction between the wet and dry seasons and high frequency of forest fires.
\subsection{Performance with Machine Learning Model}

CNNs are deep neural networks commonly used to classify images. A CNN was used to produce a continuous variable instead of classifying input. This approach allowed for data to be spatially represented in the network. The network structure of the CNN developed consists of two distinct subsections, feature learning and classification for prediction (Figure 4). Feature learning consists of convolutional and pooling layers. In between the convolutional and pooling layers, a normalisation function is applied to reduce training time and improve network accuracy. The data is normalized between 0 and 1 using feature scaling. A flattened version of the last feature learning layer is passed as input to a sequence of dense layers interposed by dropout layers. The dense layers classify features while the dropout layers disconnect a proportion of hidden nodes within the dense layers to improve network performance and generalization \citep{Srivastava2014}. The last layer is a dense layer of a single neuron followed by a linear activation. This combination outputs a continuous variable, the predicted AOD.
 
\begin{figure}[h]
\centering
\includegraphics[width=.5\textwidth]{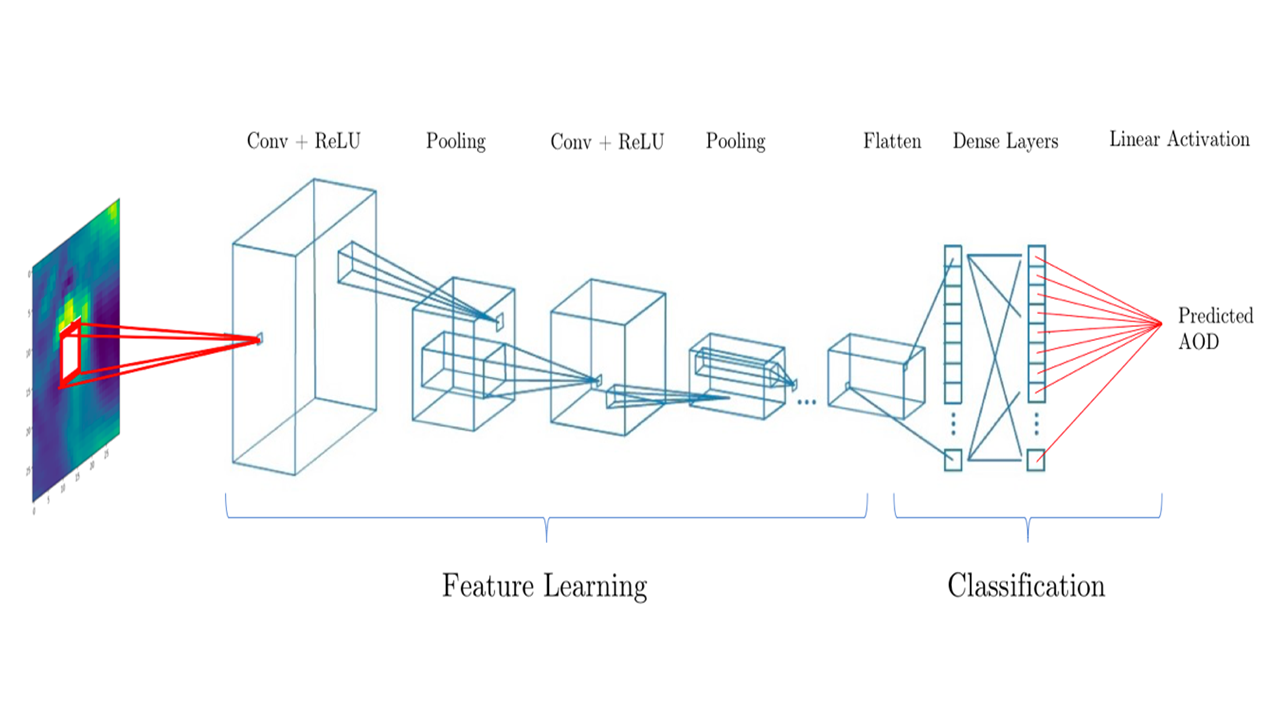}
 \caption{CNN for improving MERRA-2 AOD predictions. The full implementation used a sequence of three ‘Conv + ReLU’ and ‘Pooling’ layers followed by three ‘Dense’ layers}  
\end{figure}  

The cost function used in the CNN is a MSE function. The optimiser used to implement back propagation is the Adam optimizer. The value the learning rate is 0.003. Two distinct CNN’s were trained during the course of this analysis. The first was trained on all observations from the AERONET dataset and the second was trained on all observations where AOD was recorded to be of an extreme value. In each CNN 70\% of the data is used to train the CNN whereas the remaining 30\% is used to evaluate the performance of the network. Considering the CNN as a function, the range is the set of all observations from the AERONET dataset which meet the respective CNN criteria. Each member of the range is paired with an input. The input is a 30x30 grid of MERRA-2 data, representing a 1500km x 1875km observation area, taken at time step t-1 centered at a grid point closest to its respective AERONET observation center. The practical interpretation of this approach is to train the network to predict the AOD at an AERONET observation centre based on the MERRA-2 data from the previous day.

On a global scale, using a CNN trained model improves AOD predictions compared to MERRA-2 based physics model against the AERONET ground truth (Table 2) . 

\begin{table} [h]
  \caption{The error associated with AOD predictions decreases across all categories with the CNN}
  \label{table:lst}
  \centering
  \begin{tabular}{lllllll}
    \toprule
Model &	MERRA-2 & CNN & MERRA-2 & CNN & MERRA-2	& CNN \\
Statistic &	RMSE & RMSE & MAE	& MAE & MBE & MBE \\
\midrule
Extreme	& 1.098 &	0.775	& 0.986 &	0.436 &	-0.984 &	-0.039 \\
All	& 0.266 &	0.2419 &	0.141	& 0.135 &	-0.076 &	0.006
 \\
    \bottomrule
  \end{tabular}
\end{table}

The performance of individual sites globally with the trained CNN model predictions were further analysed. Two sites where the network performed noticeably poor were the cases of Jambi, Indonesia and Palangka Arya, Indonesia. The RMSE at these sites were 82.32 and 83.67 respectively. The next highest RMSE recorded was 11.99 at Singapore, significantly less than the cities in Indonesia. Notably, Palangka Arya and Jambi both recorded the highest variance of AOD values in the initial analysis. Across ‘all days’, the CNN performed best in locations where there were less than 5 extreme days recorded. The sites that experienced a high number of extreme days, outlined earlier improved across all performance measures except for RMSE at Kanpur, India which saw a marginal increase of 0.02. A notable outlier in the results is Hornsund, Norway. It experienced the third highest MBE and MAE amongst all sites despite only one extreme observation. A potential explanation may be due in part to its distant geographic location compared to other observation centres. Regionally, southeastern Asia performed poorly compared to most regions, however the discrepancy is not nearly as large as that of MERRA-2 in the region. Further work, could investigate different architectures, include additional variables as input features and investigate impact on regional performance.

\section*{Acknowledgements}
We thank Abhnil Prasad, Robert Taylor and Merlinde Kay from UNSW, Sydney and from CSIRO we thank, Ross Taylor, Jing Huang and Alberto Troccoli (now World Energy Meteorology Council), during Mukkavilli's PhD that partly helped motivate this work. For related machine learning discussions and feedback, we thank Gregory Dudek (McGill University) and Kris Sankaran (Mila).
\bibliography{refer}
\bibliographystyle{apa}

\end{document}